\documentclass[lettersize,journal]{IEEEtran}
\usepackage{amsmath,amssymb,amsfonts}
\usepackage[linesnumbered,ruled,vlined]{algorithm2e}
\usepackage{array}
\usepackage{textcomp}
\usepackage{stfloats}
\usepackage{url}
\usepackage{verbatim}
\usepackage{graphicx}
\usepackage{cite}
\usepackage{xcolor}
\usepackage{subcaption}
\usepackage{color,colortbl}
\usepackage{xspace}
\usepackage{hyperref}
\hypersetup{
  colorlinks   = true, 
  urlcolor     = blue, 
  linkcolor    = blue, 
  citecolor   = blue, 
  breaklinks = true 
}
\usepackage{hhline}

\definecolor{Gray}{gray}{0.9}

\newcommand{\ie}{\textit{i.e.,}\xspace}
\newcommand{\cop}{\textsc{cop}\xspace}
\newcommand{\mcts}{\textsc{mcts}\xspace}
\newcommand{\mctsplus}{\textsc{mcts}+}
\newcommand{\mctsco}{\textsc{mcts-co}\xspace}

\newcommand{\ucb}{\textsc{ucb}\xspace}
\newcommand{\uct}{\textsc{uct}\xspace}
\newcommand{\boop}{boop.\xspace}
\newcommand{\gekitai}{\(\text{Gekitai}^2\)\xspace}
\newcommand{\ghost}{\textsc{ghost}\xspace}

\newcommand{\cspmodel}[3]%
{\begin{trivlist}
  \item[]%
    \textbf{Variables:} #1\\
    \textbf{Domains:} #2\\
    \textbf{Constraints:} #3
  \end{trivlist}%
}

\newcommand{\copmodel}[4]%
{\begin{trivlist}
  \item[]%
    \textbf{Variables:} #1\\
    \textbf{Domains:} #2\\
    \textbf{Constraints:} #3\\
    \textbf{Objective function:} #4
  \end{trivlist}%
}

\newcommand{\objective}[1]%
{\begin{trivlist}
  \item[]%
    \textbf{Objective functions:} #1
  \end{trivlist}%
}

\begin{document}

\title{Injecting  Combinatorial Optimization  into \mcts:\\Application
  to the Board Game \boop}

\author{\IEEEauthorblockN{Florian Richoux}\\
\IEEEauthorblockA{\textit{AIST}, Tokyo, Japan\\
florian@richoux.fr}
}

\maketitle

\begin{abstract}
  Games,  including  abstract  board games,  constitute  a  convenient
  ground  to create,  design, and  improve  new AI  methods.  In  this
  field, Monte Carlo Tree Search is a popular algorithm family, aiming
  to  build game  trees and  explore them  efficiently.  Combinatorial
  Optimization, on  the other hand,  aims to model and  solve problems
  with an  objective to  optimize and constraints  to satisfy,  and is
  less common in Game AI.  We believe however that both methods can be
  combined efficiently,  by injecting Combinatorial  Optimization into
  Monte Carlo Tree Search to help  the tree search, leading to a novel
  combination  of these  two  techniques.  Tested  on  the board  game
  \boop, our method beats 96\% of the time the Monte Carlo Tree Search
  algorithm baseline.  We  conducted an ablation study  to isolate and
  analyze which injections and combinations of injections lead to such
  performances.  Finally,  we  opposed  our AI  method  against  human
  players on  the Board  Game Arena  platform, and  reached a  373 ELO
  rating after  51 \boop  games, with  a 69\%  win rate  and finishing
  ranked 56th worldwide on the platform over 5,316 \boop players.
\end{abstract}


\section{Introduction}\label{sec:intro}

During one  of the  51 online  games opposing our  AI agent  against a
human  player, we  were asked  in the  chat ``Why  researching new  AI
methods?''   It is  true  that  some existing  AI  methods, like  Deep
Reinforcement Learning,  would certainly  defeat any human  players at
this game, if properly trained.

Although the perspective of making an  AI agent with a deep mastery of
a game is satisfying, this is not  the reason why one does research in
Game AI.   Research is driven by  the quest to push  the boundaries of
knowledge. This  can be done  by proposing  something new. One  way to
search for  new AI methods  is to  try combining two  existing methods
that have never been combined before.

This is  what this  study aims  to do, by  combining Monte  Carlo Tree
Search and  Combinatorial Optimization  in a way  that has  been never
explored, to the  best of our knowledge. This  paper actually proposes
three possible combinations,  or to be more  specific, three different
injections of Combinatorial Optimization into Monte Carlo Tree Search,
to   improve  performances   of  the   latter.  In   particular,  such
combinations can be very profitable  on devices with limited computing
power, where only a few random playouts can be performed.

The proposed  method in  this paper  is applied  on a  recent abstract
board game called \boop (without capital  letters and with a dot.) The
main interest of this game is to be simpler than Go or Chess, but deep
enough to  offer complex  strategies.  These  characteristics motivate
its choice to be the testbed of a new method.

\section{Background}\label{sec:background}

This section  introduces the two  combined AI techniques,  Monte Carlo
Tree  Search and  Combinatorial Optimization,  as well  as \boop,  the
board game used as a testbed.

\subsection{Monte Carlo Tree Search}\label{sec:mcts}

Monte Carlo Tree Search (\mcts) is  a family of tree search algorithms
relying on the  Monte Carlo method, \ie random  samplings.

Originally  developed  for  Go~\cite{Coulom2007}, this  type  of  tree
search algorithm  has been  applied successfully  to many  other board
games  such as  Checkers, Hex  and  Backgammon, as  well as  strategy,
general  and  arcade video  games~\cite{Xu2023,Sironi2018,Pepels2014}.
\mcts has also been combined with Deep Reinforcement Learning to reach
state-of-the-art  levels at  Go,  Chess, and  Shogi~\cite{Silver2018},
among other games.

\begin{figure}[htbp]
	\centering
  \includegraphics[width=\linewidth]{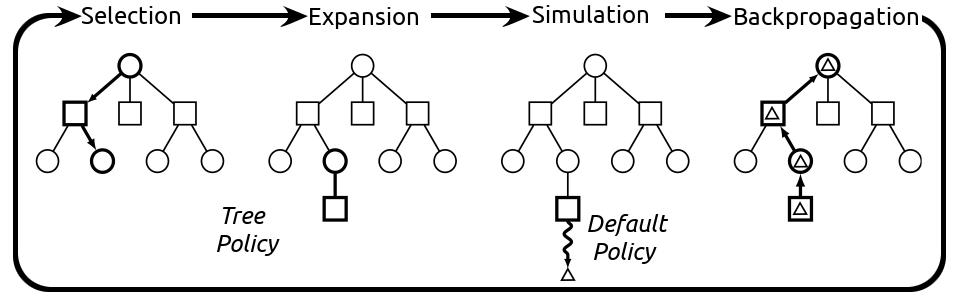}
  \caption{Steps of the Monte Carlo Tree Search.}\label{fig:mcts} 
\end{figure}

\mcts aims to build a game tree of a reasonable width, even with games
implying  a high  branch factor  like  Go, by  focussing on  promising
branches    of    the    tree.     Its    principle,    depicted    in
Figure~\ref{fig:mcts}, is  simple. It consists  in the iteration  of 4
steps:
\begin{enumerate}
\item \textbf{Selection}, where a node in the tree is chosen following
  a given \emph{Tree Policy}.
\item \textbf{Expansion}, where a new  node is inserted into the tree,
  by applying a move from the node previously selected.
\item  \textbf{Simulation},   where  moves  are   successively  chosen
  following  a \emph{Default  Policy}  (usually,  random moves)  until
  reaching a  stop criterion (usually, the end  of the game).   Such a
  series of moves is called a playout  or a rollout. In this paper, we
  call  this step  the \textbf{Playout  step}, allowing  us to  have a
  simple naming convention  for our different agents,  as explained in
  Section~\ref{sec:vs_ai}.
\item \textbf{Backpropagation}: The state reached after the simulation
  is  evaluated   by  a   function  computing   a  reward,   which  is
  backpropagated to its parent node up  to the root.  This reward will
  influence the Tree Policy during the Selection step.
\end{enumerate}
After reaching  a timeout or a  given number of iterations,  the \mcts
algorithm stops  and outputs  the move  maximizing a  given criterion,
such as the most  visited child of the root node, or  the one with the
highest  reward,  etc.   The  reader  can  refer  to  Browne  et  al's
survey~\cite{Browne2012} for further information on \mcts.

In  practice, the  Tree Policy  for Selection  is often  determined by
computing a Upper  Confidence Bound (\ucb) function~\cite{Browne2012}.
Applying \ucb on \mcts leads to the Upper Confidence bounds applied to
Trees  (\uct) algorithm~\cite{kocsis06}.   \uct is  a special  case of
\mcts.   Although the  experimental  setup in  Section~\ref{sec:vs_ai}
implements \uct,  our method  could be applied  in principle  with any
\mcts algorithm.  This  is why this paper refers to  \mcts rather than
\uct specifically.

\subsection{Combinatorial Optimization}\label{sec:co}

Combinatorial  Optimization is  the field  aiming to  model and  solve
problems  where one  must  find the  optimal  combination of  discrete
variable assignments  to maximize  or minimize an  objective function,
while satisfying  all given  constraints. Several formalisms  exist to
model  such  problems:  Linear  Programming,  Answer-Set  Programming,
etc. In this paper, we model our Combinatorial Optimization problem in
a  Constraint Programming  formalism  called Constrained  Optimization
Problems~\cite{handbookCP} (\cop).

A \cop is characterized by the quadruplet \((V,D,C,f)\), where:
\begin{itemize}
\item \(V\) is the set of decision \textbf{variables} of the problem.
\item \(D\)  is the set  of \textbf{domains}. A  domain is the  set of
  values a variable can be assigned to.
\item  \(C\)  is the  set  of  \textbf{constraints}, forbidding  some  variable
  assignment combinations.
\item \(f\) is the \textbf{objective function} to optimize.
\end{itemize}

There exist  two families of  algorithms to solve problems  modeled in
Constraint Programming:  Complete and incomplete  algorithms. Complete
algorithms cover the entire search space  by pruning it, and can proof
the   optimality   of   a   solution.    Incomplete   algorithms,   or
meta-heuristics, rely  on random moves  and heuristics to  explore the
search space. Although  such methods cannot prove the  optimality of a
solution, they are faster than complete algorithms in practice and can
tackle larger problems.

\subsection{\boop}\label{sec:pobo}

\boop is a board game created by  Scott Brady and published in 2022 by
Smirk  and Dagger  Games.  It  is the  commercial version  of \gekitai
(Gekitai squared),  released by Scott  Brady for  free in 2020  on the
website BoardGameGeek. Since  both games have exactly  the same rules,
we will refer to this game only by its commercial name \boop

\begin{figure}[htbp]
	\centering
	\begin{subfigure}[t]{0.48\linewidth}
		\centering
    \includegraphics[width=\linewidth]{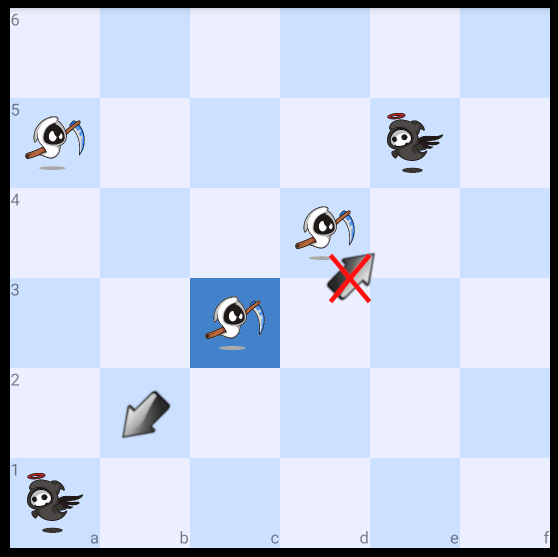}
		\caption{Push rules 1}\label{fig:push1} 
	\end{subfigure}
  \hfill
	\begin{subfigure}[t]{0.48\linewidth}
		\centering
    \includegraphics[width=\linewidth]{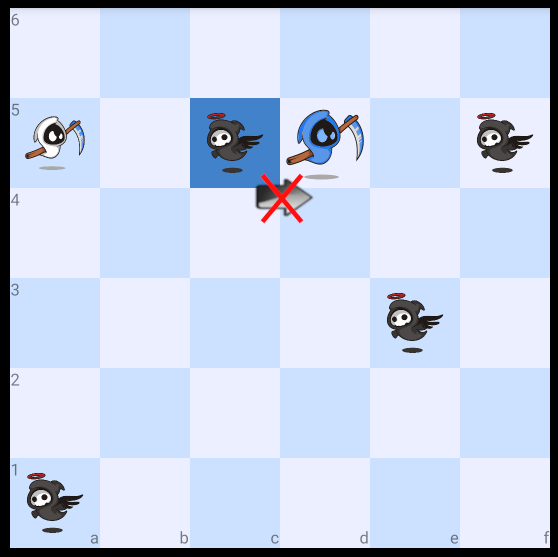}
		\caption{Push rules 2}\label{fig:push2} 
	\end{subfigure}
  \vskip\baselineskip
	\begin{subfigure}[t]{0.48\linewidth}
		\centering
    \includegraphics[width=\linewidth]{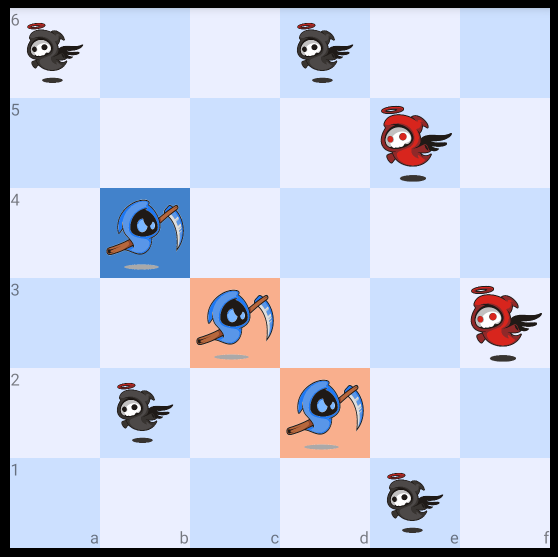}
		\caption{Victory condition 1}\label{fig:victory1} 
	\end{subfigure}
  \hfill
	\begin{subfigure}[t]{0.48\linewidth}
		\centering
    \includegraphics[width=\linewidth]{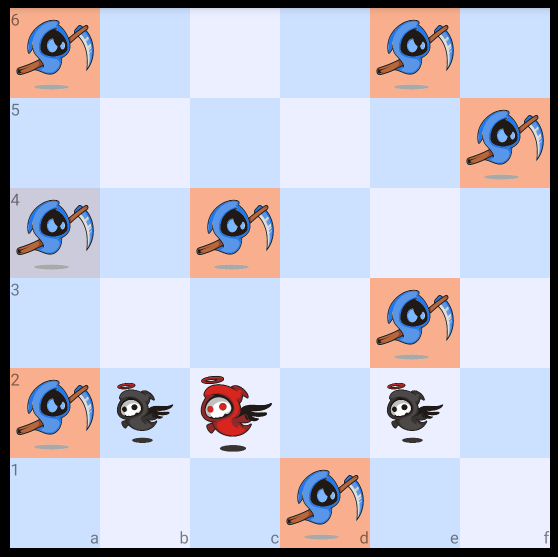}
		\caption{Victory condition 2}\label{fig:victory2} 
	\end{subfigure}
  \caption{Main rules of \boop These  images are from a \boop Android
    app we are developing.}
  \label{fig:pobo_rules}
\end{figure}

\boop is a  deterministic, fully observable, 2-player  game. The rules
are simple:  Each player has  8 small and  8 large pieces,  and starts
with a  pool of 8 small  pieces.  Players place alternately  one piece
from their pool on  a free square of the \(6 \times  6\) board. When a
piece is placed,  it pushes away all adjacent pieces  from one square,
except  if a  piece  is blocked  by another  piece,  like depicted  in
Figure~\ref{fig:push1}: A white piece has been played in c3 and pushed
away a  black piece from  b2 to  a1, but did  not push away  the white
piece in d4 because it is blocked by another piece in e5. Large pieces
can push  away any  other pieces,  but small  pieces cannot  push away
large pieces (Figure~\ref{fig:push2}).  When a  piece is pushed out of
the board, it returns into its player's pool.

When 3 pieces of a player are aligned, they are removed from the board
at  the  end of  the  player's  turn,  and  return into  the  player's
pool. Small pieces  removed that way are promoted to  large pieces. If
more than 3  pieces are aligned, the player chooses  3 adjacent pieces
to remove.   If players place  their 8 pieces  on the board,  they can
choose one  piece to remove  from the board. In  case this piece  is a
small one, it is promoted to a large piece.

A player wins the game if he or  she has 3 large pieces aligned at the
end  of his  or her  turn (Figure~\ref{fig:victory1}),  or if  8 large
pieces  are placed  on  the board  at  the end  of  the player's  turn
(Figure~\ref{fig:victory2}). There are no tied games in \boop

\section{Related work}\label{sec:related}

Many works tried to combine  \mcts and Combinatorial Optimization, and
more specifically Constraint Programming,  but always from a different
perspective than ours.

To solve a special case of the Travelling Salesman Problem encountered
in the  automotive industry,  Antuori et  al.~\cite{antuori21} combine
\mcts   and  Combinatorial   Optimization  to   improve  Combinatorial
Optimization  solvers   by  applying   \mcts  techniques   to  balance
exploration  and  exploitation  of  the  optimization  problem  search
space. This  is a fundamental difference  with our work: Where  we use
Combinatorial  Optimization to  improve  an \mcts  method, Antuori  et
al. use \mcts to improve a Combinatorial Optimization method. A common
point is  that they  replace \mcts playouts  with a  Deep-First Search
method,  when   we  replace  them   by  a  series   of  Combinatorial
Optimization problem resolutions, one for each move in the simulation.

In  the same  manner,  the Bandit  Search  for Constraint  Programming
(\textsc{bascop})  algorithm from  Loth  et  al~\cite{loth13} aims  to
adapt \mcts to the characteristics  of a Constraint Programming search
tree.  Again, the goal here is to  use \mcts to improve the search for
a combinatorial problem.  Specifically, they designed the \mcts reward
function to estimate to each couple (variable, value) a failure score,
called \textit{relative failure depth},  then exploited by the solver:
Their  algorithm  guides  the  Constraint Programming  search  in  the
neighborhood  of  the  previous  best  solution,  by  exploiting  this
relative failure depth estimated during the search space exploration.

Goffinet and Ramanujan  use \mcts to solve  the Maximum Satisfiability
Problem (a.k.a.  MaxSAT)~\cite{goffinet16}, to balance exploration and
exploitation  during the  search of  a SAT  solver.  They  propose the
\textsc{uctmaxsat}  algorithm,   where  each  node  in   the  tree  is
associated  to  a variable  in  the  SAT  formula, with  two  possible
decisions corresponding to the  variable evaluation. Playouts and leaf
node evaluations are done by two SAT stochastic local search algorithm
runs, one starting by evaluating the  expended node by true, the other
one by false.

Finally,  we  can  also  mention  Sabharwa  et  al.'s  work  to  guide
Combinatorial   Optimization  in   Mixed   Integer  Programming   with
\mcts~\cite{Sabharwal12}.

To  the  best of  our  knowledge,  all works  combining  Combinatorial
Optimization  and  \mcts methods,  like  the  related works  presented
above,  aim to  take advantage  of the  \mcts capacity  to handle  the
exploration-exploitation  dilemma to  help Combinatorial  Optimization
solvers exploring their search space.  From that perspective, the work
proposed in  this paper differs  radically from these  previous works,
combining   two  methods   the  other   way  around,   \ie  exploiting
Combinatorial Optimization capacities to  find optimal solutions under
constraints to improve a Monte Carlo Tree Search.

One can also  find many Game AI works combining  \mcts with heuristics
to improve the \mcts method.   These works include using heuristics to
bias the Tree Policy by replacing or extending the usual \ucb function
by some  heuristics for  selection~\cite{Holmgard2019,Chaslot2008}, or
using   heuristics   to   bias   the   Default   Policy   by   guiding
playouts~\cite{Finnsson2008,  Lanctot2014}.   {\'S}wiechowski  et  al.
wrote   a   good  survey   about   recent   \mcts  modifications   and
applications~\cite{Swiechowski2023}.   Our method  differs from  these
works in two aspects:
\begin{enumerate}
\item It does  not simply use a heuristics, but  solve a Combinatorial
  Optimization problem  to bias  both the  Tree and  Default Policies,
  with the advantages explained in the next paragraph.
\item The  Tree Policy is  biased without modifying nor  replacing the
  \ucb  function.   Instead,  Combinatorial Optimization  is  used  to
  narrow the  number of nodes  that can  be randomly drawn  during the
  Selection step.
\end{enumerate}

One can notice that the objective  function and the constraints of the
Combinatorial  Optimization model  could actually  be combined  into a
unique    heuristics   by    replacing   constraints    with   penalty
functions. However, there  are three main advantages to  bias the Tree
and  Default   Policies  by  modeling  and   solving  a  Combinatorial
Optimization problem rather than simply using a heuristics:
\begin{enumerate}
\item The heuristics output would  not allow to mathematically certify
  that all constraints are satisfied, unlike solving a \cop.
\item Expressing  the bias as  a \cop allows  us to take  advantage of
  solvers  containing  specific  mechanisms  to  exploit  the  problem
  structure  induced by  the  constraint network,  both with  complete
  solvers (filtering  and constraint propagation)  and meta-heuristics
  (constraint-based local search).
\item While using a heuristics, one needs to call it on every possible
  move. However,  constraint solvers  do not  explore the  entire move
  space: For instance, complete solvers prune the problem search space
  to avoid visiting subspaces where they determined that solutions are
  infeasible or  suboptimal.  Although  this feature  does not  have a
  strong impact for \boop, since the  move space of the game is small,
  this  could be  very useful  for other  games and  applications with
  significantly larger move or action spaces.
\end{enumerate}

\section{Mixing \mcts and Combinatorial Optimization}\label{sec:method}

Before describing  the Combinatorial Optimization problem,  we explain
at  the beginning  of this  section  how do  we combine  Combinatorial
Optimization and  \mcts methods. Then,  we give the intuitive  idea of
the  Combinatorial Optimization  model in  Subsection~\ref{sec:model},
followed by its formal model and some design choices.

Random playouts  are a powerful  mechanism within \mcts.   However, to
have a good estimation of the current  game state and the value of its
possible moves, one must run a significant number of playouts. This is
not always  easy to do,  depending on  the hardware: We  implemented a
vanilla \mcts method  within a \boop Android app  but quickly realized
that the number of playouts we  could run on our Android device within
a reasonable time was too small to be reliable. Within one second, the
device could only run about 80 playouts in average.

This issue  can be tackled by  replacing playouts with moves  that are
selected by  solving the Combinatorial Optimization  problem described
in  the next  Subsection~\ref{sec:model}. Algorithm~\ref{algo:playout}
illustrates how it  works. Each move of the playout  is randomly drawn
among  the moves  maximizing  the  Combinatorial Optimization  problem
(Line~4).  After  being drawn, a move  is simulated to get  a new game
state (Lines~5 and~6), and an  associated reward is computed regarding
if the move leads to a  terminal game state (Line~8) or not (Line~10).
This is  repeated until  a terminal  state is  reached or  after \(k\)
moves (while loop at Line~3).  Then, the playout stops and returns the
cumulative, normalized reward (Line~12).  The value of \(k\) we choose
in practice  is discussed in Subsection~\ref{sec:model}.   The playout
reward is  estimated by computing  a discounted sum of  the normalized
scores of  the successive \(k\)  moves, divided by \(k\).   Scores are
simply the objective function output of the Combinatorial Optimization
model and are normalized within the  range \([-1,1]\), such that -1 is
the score of a loss and 1 the score of a victory.  The discount factor
is  a parameter  \(d\) we  discuss in  Subsection~\ref{sec:model}.  We
denote by  \(a = (p,r,c)\) the  move placing a piece  of type~\(p\) on
the board at row~\(r\) and  column~\(c\).  Let \(a_1, \ldots, a_k\) be
the \(k\) moves played during a  playout.  Its playout reward \(R\) is
estimated by Equation~\ref{eq:reward}
\begin{equation}\label{eq:reward}
R = \frac{1}{k}\sum_{i=1}^{k} d^i.f(a_i)
\end{equation}
where   \(f\)  is   the  objective   function  of   the  Combinatorial
Optimization model.  Notice  that in Algorithm~\ref{algo:playout}, the
\textit{Reward}  function   on  Line   10  corresponds   to  computing
\(d^i.f(a_i)\). Since the  image of \(f\) is  \([-1,1]\), the discount
factor is  such that  \(d \leq  1\) holds,  and the  sum of  the \(k\)
products is divided  by \(k\), we have \(R \in  [-1,1]\).  The playout
reward is thus not necessarily \(0/1\) or \(-1/1\).  This is perfectly
acceptable  for  the  \ucb  function, like  described  in  Kocsis  and
Szepesvári's paper introducing the \uct algorithm: We are here dealing
with a P-game tree, that is, ``\textit{a minimax tree that is meant to
  model games where at the end of  the game the winner is decided by a
  global evaluation of  the board position where  some counting method
  is  employed}''~\cite{kocsis06},  instead  of the  classic  win/loss
evaluation.

We also  inject the same  Combinatorial Optimization problem  into the
\mcts  process  to   bias  the  Selection  and   Expansion  steps,  as
illustrated by Algorithm~\ref{algo:mcts}.  For the Selection step, the
solver  is called  to pre-select  the \(m\)  best moves  regarding the
current game state  (Line~1).  In other words, it  selects \(m\) nodes
among the  root's children.  Unselected children  are masked (Line~2),
to prevent the UCB function considering them, forcing to select one of
the  preselected children  (Line~4).   This is  analogical to  invalid
action  masking  in   Reinforcement  Learning~\cite{Huang2022},  where
invalid or poor  actions/decisions are masked mostly  at the beginning
of the learning process, to avoid confused and chaotic situations that
are usual during  the first iterations, thus  shortening the learning.
For  the  Expansion step,  the  Combinatorial  Optimization solver  is
simply called to  find what are the best moves  to play, regarding the
current  game state  and excluding  the moves  that have  been already
explored (Line~11).

Finally, we  set a  timeout of 1  second to let  our method  build and
explore the  game tree before outputting  a move to play.   The set of
moves with  the highest score/visits  ratio is computed  (Line~19) and
the algorithm returns one move randomly drawn from this set, following
a uniform distribution (Line~20).

\begin{algorithm}[htbp]
  \DontPrintSemicolon
  \KwIn{A \(node\), an integer \(k\) and a game state \(gs\)}
  \KwOut{The normalized playout score}
  iterations \(\gets\) 0\;
  score \(\gets\) 0\;
  \While{\(node\) is not terminal and iterations \(< k\)}{
    best\_move \(\gets\) Random(Solver(\(gs\)))\;
    \(node \gets\) Simulate\_move(best\_move)\;
    \(gs \gets\) Update(\(gs, node\))\;
    \If{\(node\) is terminal}{
      score \(\gets\) score + Terminal\_score(\(node\))\;
    }
    \Else{
      score \(\gets\) score + Reward(\(node\), iterations)\;
    }
    iterations \(\gets\) iterations + 1\;
  }
  \Return{score / iterations}\;
  \caption{{\sc Playout}}
  \label{algo:playout}
\end{algorithm}

\begin{algorithm}[htbp]
  \DontPrintSemicolon
  \KwIn{A game state \(gs\) and a \(root\) node}
  \KwOut{One of the best estimated moves}
  preselected\_moves \(\gets\) Solver(\(gs\))\;
  unselected\_mask \(\gets\) Childs(\(root\)) \(\setminus\) preselected\_moves\;
  \While{timeout unreached}{
    \tcp{Select a node in the tree}
    selected \(\gets\) UCT(unselected\_mask)\; 
    \(gs \gets\) Update(\(gs\), selected)\;
    \If{selected is terminal}{
      selected.visits \(\gets\) selected.visits + 1\;
      Backprop(selected.parent, selected.score)\;
      continue\;
    }
    masked\_childs \(\gets\) Childs(selected)\;
    \tcp{Expand the tree with a new node}
    expanded \(\gets\) Random(Solver(\(gs\), masked\_childs))\;
    \(gs \gets\) Update(\(gs\), expanded)\;
    expanded.parent \(\gets\) selected\;
    \If{expanded is not terminal}{
        expanded.score \(\gets\) Playout(expanded, 20, \(gs\))\;
      }
    \Else{
      expanded.score \(\gets\) Terminal\_score(expanded)\;
    }
    Backprop(selected, expanded.score)\;
  }
  best\_moves \(\gets\) Best\_ratio(preselected\_moves)\;
  \Return{Random(best\_moves)}\;
  \caption{{\sc Enhanced MCTS}}
  \label{algo:mcts}
\end{algorithm}

In summary, our method injects Combinatorial Optimization into 3 steps
of \mcts:  Just before the Selection  step (Algorithm~\ref{algo:mcts},
Line 1) and during the Expansion step (Algorithm~\ref{algo:mcts}, Line
11), to  bias to Tree Policy,  and during the Playout  step, replacing
playouts by successive  Combinatorial Optimization problem resolutions
(Algorithm~\ref{algo:playout}, Line  4), redefining a  Default Policy.
Despite these  modifications, the  resulting tree search  algorithm is
still an  \mcts algorithm because all  4 steps are applied,  and there
are still  some randomness in  the Playout step: If  the Combinatorial
Optimization  solver finds  several optimal  solutions, \ie  different
moves of  the same quality  according to the objective  function, then
one  of  these  moves  is   randomly  selected,  following  a  uniform
distribution     (Algorithm~\ref{algo:playout},     Line    4,     and
Algorithm~\ref{algo:mcts}, Line 11).  Such a situation occurs often in
a game: We ran 10 games specifically to evaluate this, and measured it
occurs in average 21,352 times per agent and per game.

It is worth  noticing that the method presented in  this paper focuses
on the ``move  decision-making'' in \boop, \ie placing a  piece on the
board. There is  actually a second type of decision  players must take
in a \boop game:  In some occasions, a player has  the choice of which
pieces to promote. In this work,  we handle this decision via a simple
heuristics favoring  taking pieces  on the border  of the  board.  All
agents  presented  in  Section~\ref{sec:vs_ai} share  this  heuristics
about ``promotion decision-making''.

The next subsection introduces  the tackled Combinatorial Optimization
problem.

\subsection{The Combinatorial Optimization model}\label{sec:model}

Before  proposing   a  formal  model  of   the  tackled  Combinatorial
Optimization problem, we first give the intuition behind it.  For node
pre-selections,  expansions,  and  playouts,  the  same  Combinatorial
Optimization problem  is solved:  Finding a  move maximizing  the game
state  score,  determined  by  a  given  heuristics  computed  by  the
objective function, such that the following constraints are satisfied:
1. The piece  we play belongs to  our pool, 2. Its position  is a free
square on the board, and 3.  The combination (piece type, position) is
not a masked move.  The two first constraints certify that the move is
valid, the last one forces finding a  move that does not belong to the
set  of masked  ones. This  is necessary  for the  node pre-selection,
where all nodes  but \(m\) are masked, but also  for the expansion, to
assure we won't regenerate an existing node.

The following  model formally  describes the problem  presented above:
\copmodel%
{\(V =  \{v_p, v_r,  v_c\}\), with \(v_p\)  the variable  deciding the
  type of piece  to play for the move, and  \(v_r, v_c\) the variables
  about the row and column number of the move position.}%
{\(D        =        \{D_{piece},       D_{position}\}\),        where
  \(D_{piece}  =  \{small, large\}\)  is  the  domain of  \(v_p\)  and
  \(D_{position}  =   \{1,\ldots,6\}\)  the  domain  of   \(v_r\)  and
  \(v_c\).}%
{\(C = \{HasPiece(v_p), FreePosition(v_r,v_c),\linebreak Unmasked(v_p,
  v_r, v_c)\}\). We formally describe  these constraints latter in the
  section.}%
{\(f(v_p,v_r,v_c)\)  is a  heuristics assigning  a score  to the  game
  state  after simulating  the  move (\(v_p,  v_r,  v_c\)). The  exact
  heuristics function formula  is rather long and not  easy to display
  clearly in  a paper, but basically,  it attributes a score  based on
  the difference  between the two players  of the number of  pieces on
  the board, on the center and  on the border, the difference of large
  pieces possessed, and if two or three pieces are aligned.  This last
  part of the score differs regarding the type of pieces composing the
  alignment. The  sum of  all these  is a number  in the  range [-MAX,
  MAX].   We divide  it by  MAX to  normalize the  outputted score  in
  [-1,1].  The  exact heuristics function  can be found in  the source
  code\footnote{\href{https://github.com/richoux/pobo/blob/21318db46e0f8fcc99d1cfaf03a9f8df7ec5d00a/app/src/main/cpp/heuristics.cpp}{heuristics.cpp}
  }.}%

The three constraints of the model can be formally described as follows:

\begin{displaymath}
  HasPiece(v_p) =
  \begin{cases}
    true & \text{if } v_p \in \text{player\_pool}\\
    false & \text{otherwise}
  \end{cases}
\end{displaymath}

\begin{displaymath}
  FreePosition(v_r,v_c) =
  \begin{cases}
    true & \text{if } (v_r,v_c) \in \text{free\_squares}\\
    false & \text{otherwise}
  \end{cases}
\end{displaymath}

\begin{displaymath}
  Unmasked(v_p, v_r, v_c) = 
  \begin{cases}
    true & \text{if } (v_p, v_r, v_c) \notin \text{mask}\\
    false & \text{otherwise}
  \end{cases}
\end{displaymath}

To be valid,  a variable assignment must be such  that all constraints
output \textit{true}.

The  model  contains  three  parameters,  already  introduced  at  the
beginning of this  section: The number \(k\) of  moves computed during
playouts,  the number  \(m\) of  pre-selected nodes  and the  discount
factor \(d\).  We did not make  an extensive parameter tuning for this
study and set their value after  some very brief trials.  An extensive
parameter tuning could probably improve  the global performance of our
method.  This  is let  as future  work. We  set \(k=20\),  \(m=5\) and
\(d=0.9\).

The C++ framework  \ghost~\cite{GHOST2016} has been used  to model and
solve   the  Combinatorial   Optimization  problem.   It  contains   a
constraint-based  local search  solver, as  well as  a backtrack-less,
complete solver since its version 3, designed to find all solutions of
the tackled problem. We use this  complete solver to find and evaluate
all possible moves in a given game state.

We can now introduce our experimental setup and results.  Two types of
experiments have been  performed: 1.  Section~\ref{sec:vs_ai} compares
our  method with  two baselines  and  with variations  of our  method,
running AI versus  AI games. 2.  Even  if our goal is not  to make the
best AI agent  playing \boop, we wanted to evaluate  its level against
human players.   To do  so, we  played the AI  agent against  28 human
players    in    51    games    on    the    platform    Board    Game
Arena\footnote{\href{https://boardgamearena.com}{https://boardgamearena.com}}.
This is detailed in Section~\ref{sec:vs_human}.

\section{AI versus AI experiments}\label{sec:vs_ai}

The  goal of  this  work  is to  improve  \mcts  methods by  injecting
Combinatorial Optimization techniques.  A  plain, vanilla \mcts method
is  therefore   a  natural  baseline.   Comparing   the  Combinatorial
Optimization-enhanced \mcts with a vanilla \mcts is easy: One just has
to disable all Combinatorial Optimization solver calls in the enhanced
\mcts to get a vanilla  \mcts implementation.  Thus, Selection is done
considering all children  of the root node, and  Expansion and Playout
are done randomly.  We did not give  our method a specific name, so we
refer to it by \mctsco in this section.

This section  also compares  \mctsco with an  agent choosing  its next
move by  only calling  the heuristics function  used in  our objective
function.   This  constitutes the  second  baseline,  to test  if  all
improvements reached by  \mctsco come from the heuristics  only, or if
it should be attributed to  the combination of \mcts and Combinatorial
Optimization.

Finally,  an ablation  study is  performed by  comparing \mctsco  with
itself when Combinatorial Optimization is  enabled or disabled for the
Selection,  the Expansion  and the  Playout steps.   We denote  agents
implementing these  modifications by  \mcts~+~the first letter  of the
concerned  steps.   For instance,  \mctsplus  SP  is the  \mcts  agent
injecting  Combinatorial Optimization  in  the  Selection and  Playout
steps. The reader can observe  that the agent \mctsco corresponds thus
to the agent \mctsplus SEP.

\subsection{Experimental setup and results}\label{sec:vs_ai_setup}

We set a timeout  of 1 second for all agents to  choose its next move,
except for the heuristics agent who  does not need any timeout because
it  does not  apply  an  iterative process:  It  calls its  heuristics
function  once on  each  possible move  and keeps  the  move with  the
highest score,  or randomly  draws one  move among  the ones  with the
highest score.

All experiments  have been  done through  a \boop  Android app  we are
developing,  running   an  Android  virtual  device   on  Linux,  thus
simulating the  limited resources  of an Android  phone compared  to a
computer.
The source  code of the  Android app,  the experimental setup  and the
results             can             be            found             at
\href{https://github.com/richoux/Pobo/releases/tag/0.6.2}{github.com/richoux/Pobo/releases/tag/0.6.2}.

Table~\ref{tab:xp} compiles results of 100  games of our \mctsco agent
against  the vanilla  \mcts agent,  all combinations  of Combinatorial
Optimization-enhanced  \mcts agents,  and the  heuristics agent.  The
\mctsco agent played  half of these games as the  first player P1, and
the other half as the second player P2.

\begin{table}[htbp]
  \caption{Number  of  victories of  our  \mctsco  agent versus  other
    agents, being the 1st player P1  or the 2nd player P2. \mctsco and
    its opponent start 50 games each.}
  \begin{center}
    \begin{tabular}{|l|c|c|c|}
      \hline
      \mctsco's opponent & \mctsco P1 & \mctsco P2 & win rate\\
      \hline
      Vanilla \mcts & 47 & 49 & 96\%\\
      \rowcolor{Gray}
      Heuristics & 30 & 50 & 80\%\\
      \hline
      \mctsplus S & 31 & 46 & 77\%\\
      \rowcolor{Gray}
      \mctsplus E & 40 & 49 & 89\%\\
      \mctsplus P & 49 & 50 & 99\%\\
      \rowcolor{Gray}
      \mctsplus SE & 24 & 42 & 66\%\\
      \mctsplus SP & 35 & 50 & 85\%\\
      \rowcolor{Gray}
      \mctsplus EP & 14 & 50 & 64\%\\
      \hline
    \end{tabular}
    \label{tab:xp}
  \end{center}
\end{table}

We see that \mctsco wins 96  games against the vanilla \mcts, over 100
games, showing  that injecting  Combinatorial Optimization  into \mcts
leads to  very significant improvements.   One could argue  that these
improvements could  be obtained  with the crafted  heuristics function
alone,   used  in   the  objective   function  of   our  Combinatorial
Optimization model. This  is not the case however,  since \mctsco also
beats  80  times over  100  games  the  heuristics agent,  our  second
baseline.  This  shows that  the gain of  performances comes  from the
combination of \mcts and  Combinatorial Optimization, rather than just
the heuristics function alone.

Games    against     different    combinations     of    Combinatorial
Optimization-enhanced \mcts  agents allow us to  estimate which parts
of our methods contribute the most to its improvements. First, one can
observe that \mctsco significantly outperforms all other Combinatorial
Optimization-enhanced   \mcts   agents.    Taken   separately,   each
Combinatorial Optimization  injection in the Selection,  the Expansion
and the Playout step does not bring much compared to the vanilla \mcts
agent, with eventually  the exception of the \mctsplus S  agent. It is
interesting to  observe that, despite  being not efficient  alone, the
Combinatorial Optimization-enhanced  Expansion step is a  key element
while combined  with either a Combinatorial  Optimization injection in
the Selection  or the  Playout step,  as illustrated  by the  win rate
difference of \mctsco versus \mctsplus SE/EP, and versus \mctsplus SP.
We can see that injecting  Combinatorial Optimization in the Expansion
step both greatly keeps up  the improvements initiated by \mctsplus S,
but  is  also  crucial for  the  Combinatorial  Optimization-enhanced
Playout step  in \mctsplus P:  Although \mctsplus P shows  the poorest
results among  all Combinatorial Optimization-enhanced  \mcts agents,
\mctsplus EP reveals itself to be the best one. We argue that the good
synergy between  the enhanced Expansion and  the enhanced Selection,
and in  particular between the  enhanced Expansion and  the enhanced
Playout, explains the excellent performance of \mctsco against our two
baseline agents.

\subsection{Investigating unbalanced results between  Player 1
  and 2}\label{sec:vs_ai_unba}

One can  observe from  Table~\ref{tab:xp} that  \mctsco's win  rate is
significantly higher  when the agent  is playing second,  loosing only
very  few games  in  that  position. There  are  two possibilities  to
explain this:  1. The  agent is  better when  playing second  for some
reasons,  2.  The  game itself  is  unbalanced and  favors the  second
player.  This  would be unusual,  since most  of the time,  the second
player is on the contrary disadvantaged in abstract games, but we know
that at  high level, some  \boop players  tend to think  that starting
second  is   actually  a  favorable   position\footnote{From  personal
  communications with  highly ranked \boop  players on the  Board Game
  Arena platform.}.

Our hypothesis is that the \mctsco agent is indeed better when playing
as the second player, and that  \boop is a correctly balanced game. To
check  our hypothesis,  we  run  mirror games,  \ie  games where  both
players are the same agent. We  made 100 mirror games with the vanilla
\mcts agent  as a control group,  to test if \boop  is a well-balanced
game  and  100  mirror  games  with  \mctsco,  the  heuristics  agent,
\mctsplus S,  \mctsplus E, and  \mctsplus P.  Results are  compiled in
Table~\ref{tab:balance}.

\begin{table}[htbp]
  \caption{Results of 100 mirror games.}
  \begin{center}
    \begin{tabular}{|l|c|c|}
      \hline
      Agents & P1 wins & P2 wins\\
      \hline
      Vanilla \mcts & 47 & 53\\
      \rowcolor{Gray}
      \mctsco & 7 & 93\\
      Heuristics & 60 & 40\\
      \rowcolor{Gray}
      \mctsplus S & 46 & 54 \\
      \mctsplus E & 46 & 54\\
      \rowcolor{Gray}
      \mctsplus P & 14 & 86\\
      \hline
    \end{tabular}
    \label{tab:balance}
  \end{center}
\end{table}

Games of the vanilla \mcts agents  indicate that there might have some
slight advantage for  the second player, although  statistics over 100
games are not enough to  draw solid conclusions.  The balance question
would deserve  a deeper  investigation.  Moreover,  balancing abstract
games is notoriously difficult: Chess  is considered to be a correctly
balanced game, however White playing  first has greater chance to win:
Chessgames.com 2023  statistics indicates  that White wins  57.13\% of
games            not            finishing            with            a
draw\footnote{\href{https://www.chessgames.com/chessstats.html}{https://www.chessgames.com/chessstats.html}}.
With 53\%  of win rates  for the  vanilla \mcts agent  player starting
second, it  is fair  to consider  that \boop  is a  correctly balanced
game. This  is also  confirmed by  mirror games  with the  \mctsplus S
agent and the \mctsplus E agent.

Mirror games  of \mctsco indisputably  shows that the agent  is better
when  playing second.   We  first thought  this was  only  due to  the
heuristics function used  by the objective function in  our model, but
this is  in contradiction with  the results of the  heuristics agent's
mirror games. Nevertheless, our heuristics  function still seems to be
the culprit, but in a more complex  way, when it is used repeatedly to
anticipate  the next  moves, like  in the  \mctsplus P  agent, and  of
course like in \mctsco.

Indeed, playing first at \boop  requires a particular attention on the
positioning of  our own  pieces and  on decisions  to make:  The first
player needs  to ``attack'' at the  right moment, \ie trying  to build
alignments of its pieces, neither too  early nor too late in the game,
whereas the role  of the second player consists more  to ``defend'' in
early game, trying  to break down the first  player's formations. This
may be  an easier role to  manage, and that is  currently better taken
into account by our heuristics function. This bias is amplified by the
successive calls  of the objective  function (and then  the heuristics
function)  in  Combinatorial   Optimization-enhanced  Playout  steps,
leading to  a stronger  defense and  a weaker  attack as  well. Taking
better  early  game  decisions  for  the first  player  would  ask  an
extensive specialization of the heuristics  function, which is not the
goal of our work in this paper.

\section{AI versus Human experiment}\label{sec:vs_human}

To have a first estimation of  the \mctsco agent's level against human
players,  we ask  the permission  to the  Board Game Arena platform  for
creating       an       account      specifically       for       this
agent\footnote{\href{https://boardgamearena.com/player?id=95213950}{https://boardgamearena.com/player?id=95213950}}.

We deployed the following process: After making an announcement on the
Board Game Arena  forum about our AI agent account,  as the Board Game
Arena platform recommended us to do,  we created \boop games from this
account and waited for someone to join.  We never joined games created
by other players.  At the beginning of  the game, we used  the chat to
warn the  opponent that he  or she is  playing against an  AI, telling
that it is possible to cancel the  game without any penalties if he or
she is not comfortable with that. Therefore, all opponents were warned
they  were playing  against  an AI  agent, and  all  games taken  into
account for the experiment are games where the opponent agrees to play
against the AI.  To ensure this, games have been done ``manually'': We
played from  the AI  agent account  on a computer  next to  an Android
tablet  running   the  \boop  app  implementing   \mctsco.   Then,  we
reproduced each move from the Board Game Arena opponent on the Android
tablet as  a human player,  and played on  Board Game Arena  each move
decided by  the \mctsco  agent in the  app.  This way,  we acted  as a
human operator reproducing moves from  Board Game Arena to the Android
app and from  the Android app to Board Game  Arena, responding also to
eventual questions from opponents on Board Game Arena.

Board Game Arena is implementing its  own ELO rating, which should not
be directly  compared to  Chess ELO rating  for instance.   Board Game
Arena  attributes a  rank to  players according  to their  ELO rating:
Beginner (0  ELO points),  Apprentice (1-99), Average  (100-199), Good
(200-299), Strong (300-499), Expert  (500-699), and Master (700+).  At
the  time this  experiment was  conducted, there  were 6  \boop Master
players  only on  the  Board  Game Arena  platform,  over 5,316  \boop
players.

\begin{figure}[htbp]
	\centering
  \includegraphics[width=\linewidth]{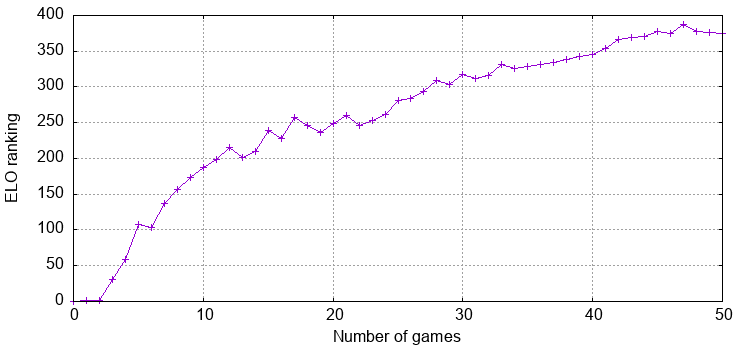}
  \caption{ELO rating of \mctsco on Board Game Arena against human players.}\label{fig:elo} 
\end{figure}

Our agent played 51 games against 28 players with ELO points from 0 to
865,  between the  13th of  December, 2023  and the  10th of  January,
2024. It won  35 games (69\% of  win rate), and finished  with 373 ELO
points  (Strong rank),  ranked  56th worldwide  on  the platform.   It
reached  the Strong  rank after  its 28th  game.  Figure~\ref{fig:elo}
illustrates the progression  of our agent's ELO  points.  Although its
ELO points  evolution looks rapid  at first  glance, it should  not be
directly compared with  the evolutions of human players  on Board Game
Arena, since  many players  are likely  to discover  the game  on this
platform and  then start  from a completely  beginner level,  when our
agent played its first games at full strength.

Considering  Figure~\ref{fig:elo},  one  could think  that  the  agent
reached its  top performances  against humans  players, since  the ELO
points curve seems to converge just  below 400 points. We do not think
it is  the case, though,  and believe that  it could go  further, even
maybe reaching  the bar of 500  ELO points. The three  last games were
played against its strongest opponent, a  Master player with a 865 ELO
rating, ranked 3rd worldwide at that time.  Our agent lost these three
games and that is what makes the curve flat at the end.

\section{Conclusion and perspectives}\label{sec:conclusion}

We presented in this paper three different injections of Combinatorial
Optimization into  Monte Carlo  Tree Search  (\mcts): Just  before the
Selection  step, during  the Expansion  step, and  during the  Playout
step.   While   previous  works   combine  \mcts   with  Combinatorial
Optimization  solvers  to  improve  them,   this  is  the  first  time
Combinatorial Optimization has been combined with \mcts to improve the
latter, to the best of  our knowledge.  Experimental results show that
a   Combinatorial   Optimization-enhanced  \mcts   algorithm   greatly
outperforms the vanilla \mcts algorithm:  In the board game \boop, our
methods wins 96\% of its games  against vanilla \mcts, and 80\% of its
games against  an heuristics-based  agent, the  second baseline,  on a
virtual device  simulating the limited computing  resources an Android
device may  offer, compared to  a personal  computer.  We also  did an
ablation study allowing us to analyze which Combinatorial Optimization
injections are essentials for  reaching these performances.  From this
ablation study, we conclude  that injecting Combinatorial Optimization
into the  Expansion Step is  the key  stone of our  method, performing
poorly  alone   but  extremely  well   while  combined  with   both  a
Combinatorial  Optimization  injection  into  the  Selection  and  the
Playout steps.

In parallel of  AI versus AI experiments, we also  ask the opportunity
to  the Board  Game Arena  platform to  let our  AI agent  plays \boop
against human players,  allowing us to have a rough  estimation of its
ELO rating.   Our agent plays 51  games against 28 opponents  of very
different  skills, winning  69\% of  its games  (35 wins,  16 losses),
finishing with a  373 ELO rating (in the ``Strong  players'' class on
Board Game Arena) and ranked 56th worldwide on the platform over 5,316
\boop players.

Apart from trivial  improvements we could bring to our  method and its
implementation, such as the tuning of its 3 parameters, an interesting
perspective could be modeling and solving a Combinatorial Optimization
problem   going  beyond   one-stage  decision-making:   So  far,   the
Combinatorial Optimization problem we solve aims to find the best move
in the current  situation, and the combinatorial part  of this problem
is certainly under-exploited for the  solver we use.  Tackling k-stage
decision-makings,  \ie   deciding  the  move  after   considering  k-1
successive  moves,   would  constitute   a  great  challenge   from  a
combinatorial  point of  view,  for instance  by  certifying that  the
opponent does  not have any direct  winning moves next after  our move
(unless all our moves are losing moves).

\section*{Acknowledgment}

We would like to thank the  Board Game Arena platform for accepting to
let us  creating an account and  playing games with our  \mctsco agent
against human players, giving us the opportunity to estimate its Board
Game Arena  ELO rating.  We  also thank  all players from  Board Game
Arena who accepted to play one or several games against our AI agent.

\vspace{0.5cm}

\noindent%
Distributed under  a Creative Commons Attribution 4.0 International
license\footnote{\href{https://creativecommons.org/licenses/by/4.0/}{https://creativecommons.org/licenses/by/4.0/}} (CC-BY 4.0).
\begin{center}
  \includegraphics[width=0.5\linewidth]{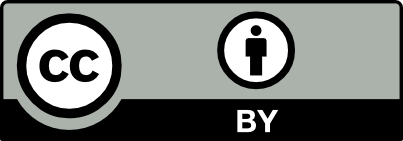}
\end{center}

Figure~\ref{fig:mcts}  is a  modification of  a figure  from Wikimedia
Commons under CC-BY-SA license. The source can be found at \href{https://en.wikipedia.org/wiki/File:MCTS-diagram.svg}{en.wikipedia.org/wiki/File:MCTS-diagram.svg}.

\bibliographystyle{IEEEtran}
\bibliography{IEEEfull,pobo}

\end{document}